\documentclass{article}

\usepackage{arxiv}

\usepackage[utf8]{inputenc} 
\usepackage[T1]{fontenc}    
\usepackage{hyperref}       
\usepackage{url}            
\usepackage{booktabs}       
\usepackage{amsfonts}       
\usepackage{nicefrac}       
\usepackage{microtype}      
\usepackage{lipsum}
\usepackage{graphicx}
\usepackage{caption}
\graphicspath{ {./images/} }

\title{MEL: Legal Spanish language model}

\author{
  David Betancur Sánchez\\
  Instituto de Ingeniería del Conocimiento (IIC) \\
\texttt{david.betancur@iic.uam.es} \\
  \And
  Nuria Aldama García\\
  Instituto de Ingeniería del Conocimiento (IIC) \\
\texttt{nuria.aldama@iic.uam.es} \\
  \And
  Álvaro Barbero Jiménez\\
  Instituto de Ingeniería del Conocimiento (IIC) \\
\texttt{alvaro.barbero@iic.uam.es} \\
  \And
  Marta Guerrero Nieto\\
  Instituto de Ingeniería del Conocimiento (IIC) \\
\texttt{marta.guerrero@iic.uam.es} \\
  \And
  Patricia Marsà Morales\\
  Instituto de Ingeniería del Conocimiento (IIC) \\
\texttt{patricia.marsa@iic.uam.es} \\
  \And
  Nicolás Serrano Salas\\
  Instituto de Ingeniería del Conocimiento (IIC) \\
\texttt{nicolas.serrano@iic.uam.es} \\
  \And
  Carlos García Hernán\\
  Instituto de Ingeniería del Conocimiento (IIC) \\
\texttt{carlos.garcia@iic.uam.es} \\
 \And
  Pablo Haya Coll\\
  Instituto de Ingeniería del Conocimiento (IIC) \\
\texttt{pablo.haya@iic.uam.es} \\
  \And
  Elena Montiel Ponsoda\\
  Ontology Engineering Group \\
  Universidad Politécnica de Madrid \\
\texttt{elena.montiel@upm.es} \\
  \And
  Pablo Calleja Ibáñez\\
  Ontology Engineering Group \\
  Universidad Politécnica de Madrid \\
\texttt{pcalleja@fi.upm.es} \\
}

\begin{document}
\maketitle
\begin{abstract}
Legal texts, characterized by complex and specialized terminology, present a significant challenge for Language Models. Adding an underrepresented language, such as Spanish, to the mix makes it even more challenging. While pre-trained models like XLM-RoBERTa have shown capabilities in handling multilingual corpora, their performance on domain specific documents remains underexplored. This paper presents the development and evaluation of MEL, a legal language model based on XLM-RoBERTa-large, fine-tuned on legal documents such as BOE (Boletín Oficial del Estado, the Spanish oficial report of laws) and congress texts. We detail the data collection, processing, training, and evaluation processes. Evaluation benchmarks show a significant improvement over baseline models in understanding the legal Spanish language. We also present case studies demonstrating the model's application to new legal texts, highlighting its potential to perform top results over different NLP tasks.
\end{abstract}


\section{Introduction}
\paragraph{Background.}
The legal domain presents challenges for Natural Language Processing (NLP) solutions due to the highly specialized nature of legal texts, specific terminology, precise references to laws, and archaic language that often differs from everyday communication. 
Recent advances in NLP, particularly since the introduction of the transformer architecture \cite{vaswani2023attentionneed} and transformer-based models such as BERT \cite{bert} and XLM-RoBERTa \cite{xlmroberta}, have demonstrated remarkable abilities to understand human language in a wide range of domains and languages, including Spanish \cite{wu2019betobentzbecassurprising}. However, while these models perform well on general text, their effectiveness in a highly specific domain such as the legal domain, especially those written in Spanish, is still underexplored.
Some multilingual legal corpora like the MultiLegalPile \cite{niklaus2024multilegalpile689gbmultilinguallegal} are available but, as it's multilingual, the focus on specific languages is not usually guaranteed and can fall down on size or even quality of cleaning. Proprietary models like legal-xlm-roberta-large \cite{huggingfaceJoelniklauslegalxlmrobertalargeHugging} use this corpus but still achieve similar results to the base model (xlm-roberta-large) at Spanish tasks evaluated in our benchmark at section \ref{Benchmark Results}.

\paragraph{Motivation.}
In recent years, the application of NLP models to domain-specific tasks has gained increasing attention, particularly in areas such as law and medicine. Legal professionals frequently work with large volumes of complex documents and the manual analysis of these documents is time consuming, prone to human error, and often requires expert knowledge. As a result, there is a growing demand for automated tools that can assist in processing and extracting meaningful information from these legal texts. However, it is important to note that Spanish is usually underrepresented in the training corpora of large language models, creating a challenge for developing NLP tools for Spanish legal texts.

\paragraph{Objective.}
This paper aims to present MEL (Modelo de Español Legal), a legal language model based on XLM-RoBERTa-large for understanding Spanish legal texts. The model will be trained on a curated dataset of Spanish legal documents and evaluated against baseline models to measure its performance in various NLP tasks, such as legal classification and named entity recognition (NER). The goal is to demonstrate the model's ability to outperform existing multilingual models in accurately capturing the features of legal Spanish, offering a valuable tool for legal professionals and NLP researchers working on these domain specific tasks.

\paragraph{Content.}
In this paper, first we discuss related work on domain-specific language models, multilingual and underrepresented languages, and legal Spanish models and their challenges. After that we will go through the methodology used to obtain the training corpus, train and benchmark our model. Finally, the results of both the training and benchmark against another models.

\section{Related Work}

\paragraph{Encoder-only vs LLMs.}
Despite the recent hype with generative LLMs, encoder-only models remain widely used in a variety of non-generative downstream applications \cite{warner2024smarterbetterfasterlonger}. They offer a great trade-off between quality and size, making it's inference requirements much better than the ones offered in LLMs. Also, on tasks such as NER and information extraction, generative models still lack intuition about correct positioning the entities, shown on the experiments performed on \cite{subies2023surveyspanishclinicallanguage}.

\paragraph{Domain-Specific Language Models.}
The adaptation of pre-trained language models for specific domains has become a prominent area of research in NLP. Models like BioBERT \cite{biobert} and SciBERT \cite{scibert}, fine-tuned for biomedical and scientific texts, respectively, have shown significant performance improvements over general-purpose models in domain-specific tasks. Legal NLP has followed a similar trajectory, with models like Legal-BERT \cite{chalkidis-etal-2020-legal} and CaseLawBERT \cite{caselawbert} fine-tuned on legal corpora, showing superior performance in tasks such as contract analysis, court judgment summarization, and legal question answering. These works underscore the importance of domain-specific fine-tuning for achieving state-of-the-art results.

\paragraph{Multilingual and open-access data underrepresented Languages.}
Multilingual models like XLM-RoBERTa and mBERT have significantly improved text understanding across multiple languages, including underrepresented ones like Spanish. In some cases, they even outperform monolingual models, particularly because many existing monolingual models for Spanish are older. However, their effectiveness can still vary depending on the specific domain and task at hand. \cite{PLN6487}

\paragraph{Legal NLP Spanish Models.}
Recent work combines both domain adaptation and multilinguality. A notable example is legal-xlm-roberta-large \cite{joelniklaus2023legalxlmr}, a multilingual transformer pre-trained on the Multi-Legal-Pile \cite{Niklaus2023MultiLegalPileA6}, a comprehensive multilingual dataset derived from diverse legal sources across 24 languages. This pre-training approach has demonstrated promising results in legal document classification and NER tasks, showcasing the benefits of incorporating domain-specific and multilingual corpora. However, despite these advancements, the application of such models to Spanish legal texts, specifically using datasets like BOE and parliamentary records, remains underexplored.

\paragraph{Challenges in Legal Language.}
 Despite the above advancements, there remains a gap in models trained specifically on Spanish legal texts. These texts often contain archaic and highly specialized terms and structures that differ significantly from everyday language. Additionally, legal texts require precise interpretation, as minor differences in documents can alter the meaning of a law or legal ruling. This complexity, added to the particularity of underrepresented languages like Spanish, and the errors presented on OCR transcription, constitute a new layer of difficulty when processing legal texts. Our research seeks to address these challenges by pre-training a multilingual transformer model on Spanish legal corpora, enabling better understanding and application of legal language in NLP tasks such as text classification and NER.

\section{Methodology}
\subsection{Corpus collection}
To pre-train MEL, we curated a specialized corpus consisting of various legal documents, including the Boletín Oficial del Estado (BOE), parliamentary proceedings, legal statutes and court rulings \cite{datos_congreso}. These documents were collected from publicly available Spanish repositories and legal databases.

The data extracted from the official journals, referred to as Boletines Oficiales, were sourced at the notice and rule level wherever possible. These journals were retrieved as PDF files from the websites of the Spanish State and its 20 autonomous communities. Given the diverse formats and structures of these sources, 20 specific web trackers were developed, one for each autonomous community. In certain cases, it was not possible to extract individual notices or rules directly as PDFs; for example, in the autonomous city of Ceuta, entire journal PDFs had to be downloaded, or the transcriptions available on the bulletin websites were used instead. Furthermore, in bilingual autonomous communities, the bulletins were downloaded considering language distinctions. It was observed that some notices were published in one language, with only the titles translated into the other, necessitating careful handling to prevent introducing unintended multilingual data into the corpus.

The PDF files adhere to Directive (EU) 2016/2102 \cite{pdfdirective}, which ensures accessible protocols, facilitating accurate transcriptions. To ensure the extraction of relevant information, extraneous elements such as margins and stamps were cropped, isolating the content related to notices and laws. Furthermore, a comprehensive evaluation of various transcription tools and Optical Character Recognition (OCR) systems was conducted. The tools compared included OCRmyPDF, pdftotext, PDFPlumber and PyPDF. Following this analisys, PDFPlumber was identified as the optimal solution for this specific use case \cite{Singer-Vine_pdfplumber_2024}.

\subsection{Corpus processing}
The corpus underwent a preprocessing state to standardize and clean the data. This process involved:
\begin{itemize}
\item \textbf{Language Cleaning}: We ensured that all documents were written in Spanish, removing any non-Spanish content to maintain linguistic consistency throughout the corpus. This was performed using Facebooks fasttext library \cite{Facebookresearch} selecting the texts that were mainly in Spanish with a confidence higher than 95\%.
\item \textbf{Unwanted characters Cleaning}: The text was cleaned to remove unwanted characters such as extra whitespaces and breaklines. This helped to standardize the corpus and minimize noise during model training.
\item \textbf{Chunking}: Current model architectures have a predefined context window. In our case the window was of 512 tokens. Texts were split in sentences and appended until reaching a maximum of 512 tokens to avoid unnecessary truncation in the training.

\item \textbf{Size}: The final corpus after tokenization and chunking consists of 5\,520\,000 texts, resulting on a dataset of 92.7 GB.
\end{itemize}

\subsection{Model training}
\begin{figure}[h]
    \centering
    \includegraphics[width=0.49\textwidth]{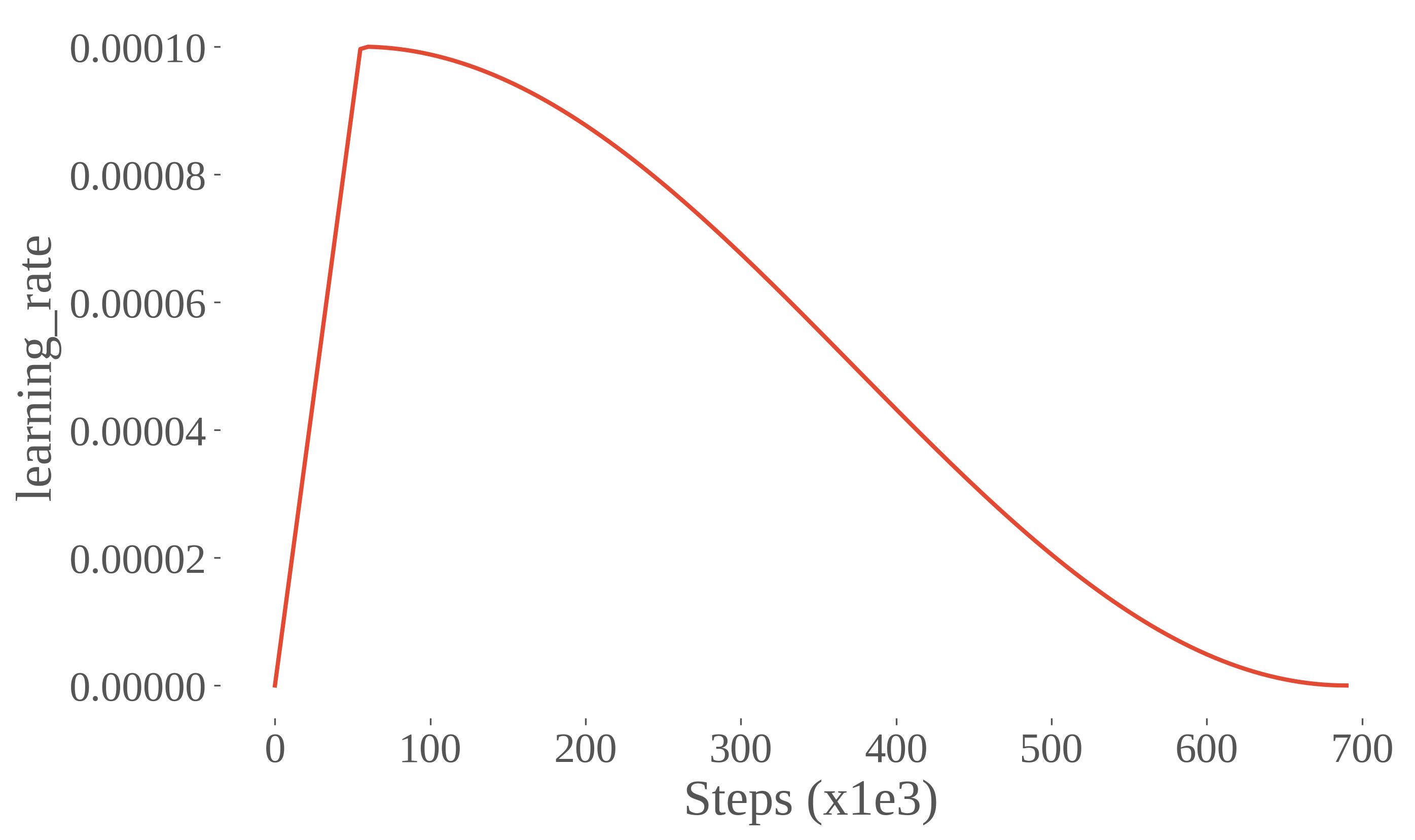}
    \caption{Learning rate}
    \label{fig:train_learning_rate_image}
\end{figure}

A domain adaptation was performed using the Huggingface Transformers Trainer \cite{huggingfaceTrainer}. The following are   some important aspects relevant to the training of the model:
\begin{itemize}
\item \textbf{Training setup and architecture}: The base model used for this task was XLM-RoBERTa-large, a transformer model pre-trained on multilingual data. Different studies such as \cite{PLN6487} demonstrate that XLM-RoBERTa-large is the best model across the encoder-only available, scoring even better than the much newer mDeBERTa.

In this phase, we extended the pre-training using a large corpus on Spanish legal texts, rather than finetuning the model for specific tasks. The architecture remained consistent with XLM-RoBERTa-large, using the same transformer layers, while the training objective focused on masked language modeling (MLM) to reinforce context-based learning in the legal domain. The whole word mask strategy was maintained with random selection of whole words replacing 80\% by [MASK], 10\% by a random token and 10\% left unchanged.

\item \textbf{Training Data}: The dataset for the continuation of pre-training included a vast collection of Spanish legal texts from BOE, parliamentary transcripts, court, rulings and other legislative documents. Processing of the data is mentioned in the section above. A random selection of 100\,000 texts was used for validation of the training.

\item \textbf{Hyperparameters}: 
\begin{itemize}

\item \textbf{learning rate}: $1 \times 10^{-4}$, selected after the one used in the base model.

\item \textbf{Optimizer}: The AdamW optimizer was used, with $\beta_1 = 0.9$, $\beta_2 = 0.98$, and $\epsilon = 1 \times 10^{-6}$. These values were selected after the good performance on the base model.

\item \textbf{Batch size and gradient accumulation}: Batch size of 16 with a gradient accumulation of 4. Effective batch size of 64, helping the model process large chunks of data while keeping memory usage manageable. This is the max batch size that fits in the Hardware used, an NVIDIA A100 with a memory of 80GB.
\item \textbf{Epochs}: 2
\item \textbf{Warmup}: 0.08. Meaning 8\,\% of the total steps were used to gradually increase the learning rate, as seen on Figure \ref{fig:train_learning_rate_image}, helping the model prevent from making abrupt weight updates in early training.
\item \textbf{Weight decay}: 0.01
\item \textbf{Scheduler}: A cosine learning rate scheduler was used. Shown on Figure \ref{fig:train_learning_rate_image}.

\end{itemize}

\item \textbf{Hardware}: Given the scale of the pre-training (large batch sizes and complexity of large models), high performance hardware with substantial GPU memory and processing power is required. In this case a NVIDIA A100 80GB PCIe was used to complete the training. 

\item \textbf{Compute time}: The continuation of the pre-training took approximately 13.9 days with each epoch lasting between 6 and 7 days. Early stopping was not applied since the objective was to pretrain the model over a fixed number of epochs. 
\end{itemize}

\subsection{Evaluation} \label{Evaluation}
The evaluation of MEL was challenging because suitable legal text datasets for machine learning tasks are notoriously scarce, particularly for non-English languages such as Spanish. The lack of labeled legal data is a significant barrier to the development of robust fine-tuned models. Legal documents often require specialized annotation by domain expert linguists, making the labeling process both time-consuming and costly.
For MEL, two benchmarks were performed both on a public and a private dataset:

\begin{itemize}
\item \textbf{Multieurlex Spanish multilabel dataset} \cite{chalkidis2021multieurlexmultilingualmultilabel}: This dataset comprises 65000 EU laws annotated  with EUROVOC concepts by the EU Publication Office. Each law is associated with one or more EUROVOC labels, where each label corresponds to a specific legal topic. For example, labels might include descriptors such as [60, agri-foodstuffs], [6006, plant product], [1115, fruit]. This descriptors, and their associated IDs, are available in 23 official EU languages, including Spanish.

\item \textbf{Multiclass classification on private dataset}: This task was conducted on a private legal corpus with texts labeled on different classes. The corpus contains a train set with 2389 texts and a test with 892 texts. There is a total of 9 different classes for the model to predict on.

\end{itemize}

The models used for the benchmark to compare against MEL include:

\begin{itemize}
\item \textbf{xlm-roberta-large} \cite{xlmroberta}: FacebookAI's multilingual version of RoBERTa. It is pre-trained on 2.5 TB of filtered CommonCrawl data containing 100 languages. 
\item \textbf{PlanTL-GOB-ES/RoBERTalex} \cite{huggingfacePlanTLGOBESRoBERTalexHugging}: PlanTL (Plan de impulso de las Tecnologías del Lenguaje) transformer-based masked language model for the Spanish language. It is based on the RoBERTa base model and has been pre-trained using a large Spanish Legal Domain Corpora, with a total of 8.9 GB of text.
\item \textbf{joelniklaus/legal-xlm-roberta-large} \cite{huggingfaceJoelniklauslegalxlmrobertalargeHugging}: Multilingual model pretrained on legal data. It is based on XLM-R large. For pre-training, Multi Legal Pile \cite{Niklaus2023MultiLegalPileA6} was used, a multilingual dataset from various legal sources covering 24 languages.
\end{itemize}

\section{Results}
\subsection{Training Performance}
\begin{figure}[ht]
  \begin{minipage}[b]{0.49\textwidth}
    \includegraphics[width=\textwidth]{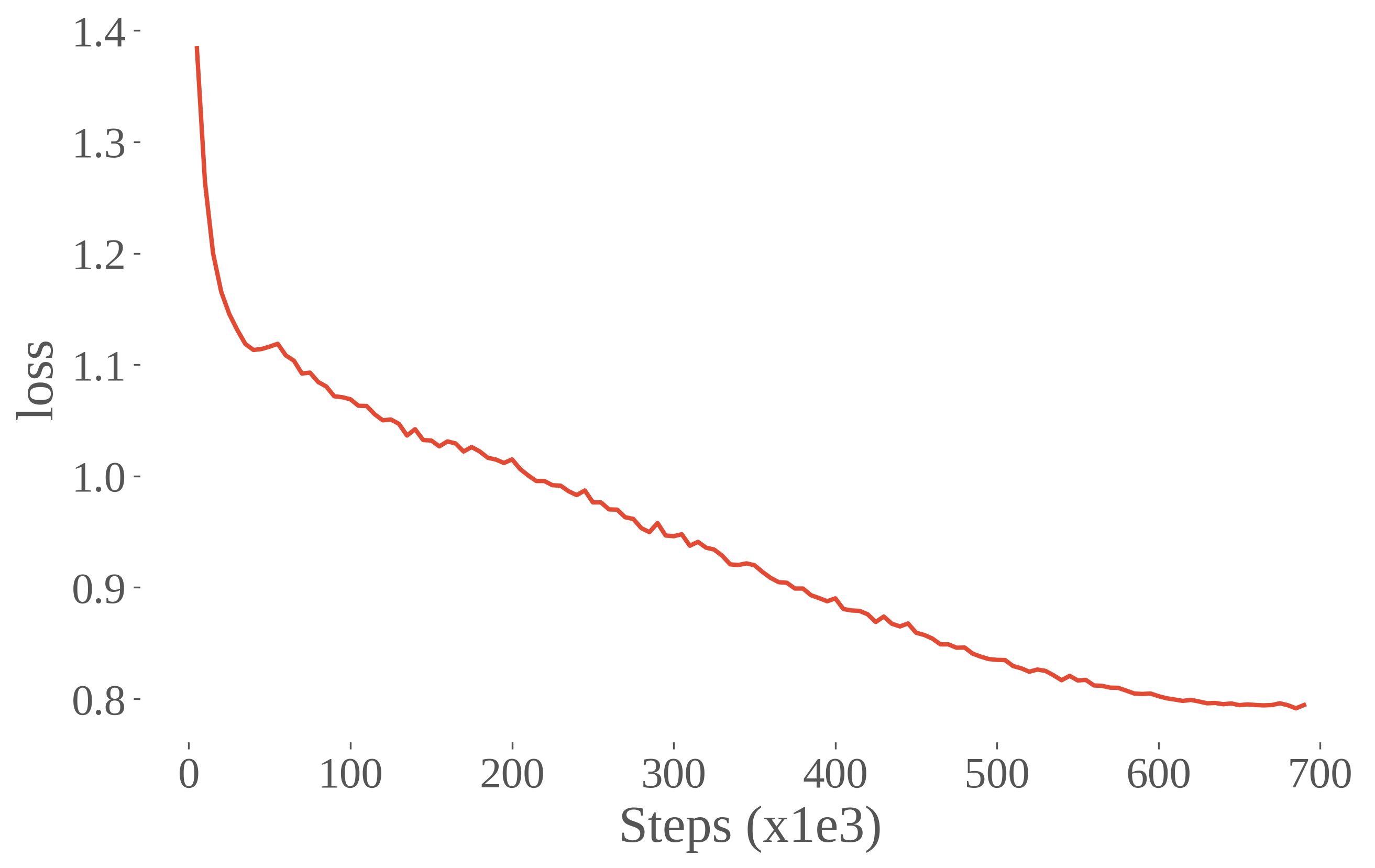}
    \caption{Evaluation Loss}
    \label{fig:eval_loss_image}
  \end{minipage}
  \hfill
  \begin{minipage}[b]{0.49\textwidth}
    \includegraphics[width=\textwidth]{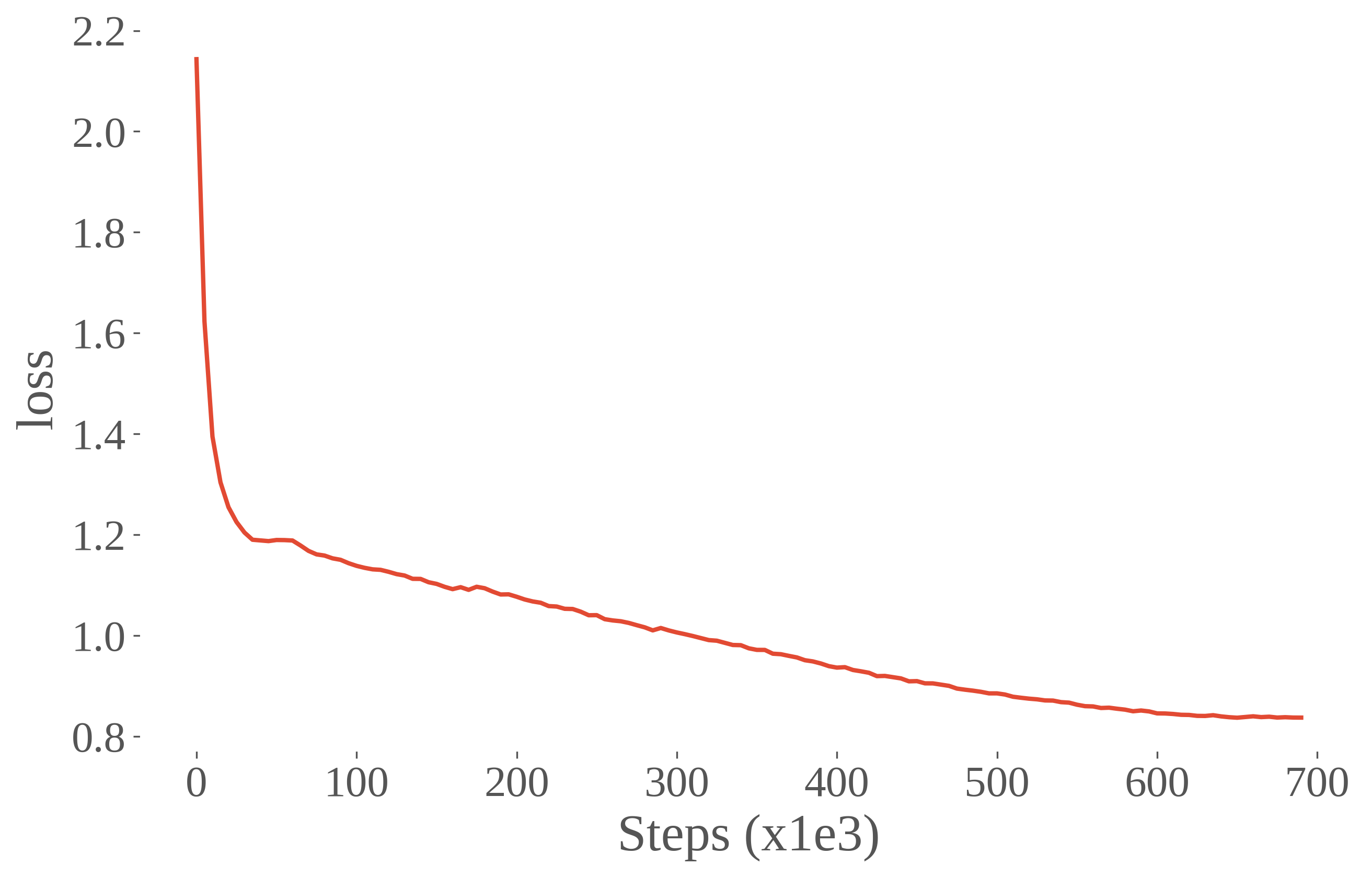}
    \caption{Training Loss}
    \label{fig:train_loss_image}
  \end{minipage}
\end{figure}

The training phase demonstrated consistent improvements in the training steps, as shown in Figures \ref{fig:train_loss_image} and \ref{fig:eval_loss_image}. The training loss shows a steady decline, indicating that the model effectively learned to fit the training data. However, the evaluation loss decreased also, reflecting the model's capacity to generalize and adapt to the features of the legal Spanish texts.
Some aspects of the training results include:

\begin{itemize}
\item \textbf{First steps of training}: During the initial steps of the training, both train and eval loss showed rapid declines, indicating that the model quickly captured fundamental patterns in the data. Warmup steps, used as in the base model training strategy, seemed to allow gradual adaptation, preserving the useful representations learned during pretrained while allowing fine-tuning for the new domain.
\item \textbf{Final steps of training}: In the later stages, the losses converged more gradually guaranteeing stability of the training.
\item \textbf{Overfitting check}: The similar steep or parallel trend of the training and evaluation loss throughout training suggests minimal overfitting, further emphasizing the generalizability of the model.
\end{itemize}

\subsection{Benchmark Results} \label{Benchmark Results}
As mentioned in section \ref{Evaluation}, to evaluate the pre-trained model, we conducted a comprehensive benchmarking analysis on multilabel and multiclass classification using both multieurlex and a private legal corpus. The F1 score over the finetuning epochs was saved and plotted of Figures \ref{fig:results_eurlex_image} and \ref{fig:legal_classification_results_image}. Sometimes, models tend to understand the tasks on similar proportions and end having similar scores, but at different rates. That is why we considered important to evaluate the learning speed of the models represented by the area under the curve of the F1 vs Epochs plot. These results can be visualized on Tables \ref{table:results_eurlex_table} and \ref{table:legal_classification_results_table}.

\begin{itemize}
\item \textbf{Multieurlex Spanish multilabel dataset}:
On the multieurlex dataset, our model seems to excel, achieving higher F1 scores at a higher learning speed than the rest of the models. It also seems that our model achieves good results on early epochs, meaning that the model can achieve good results with low or nearly-none finetuning.

\begin{figure}[ht]
  \begin{minipage}[b]{0.49\textwidth}
    \includegraphics[width=\textwidth]{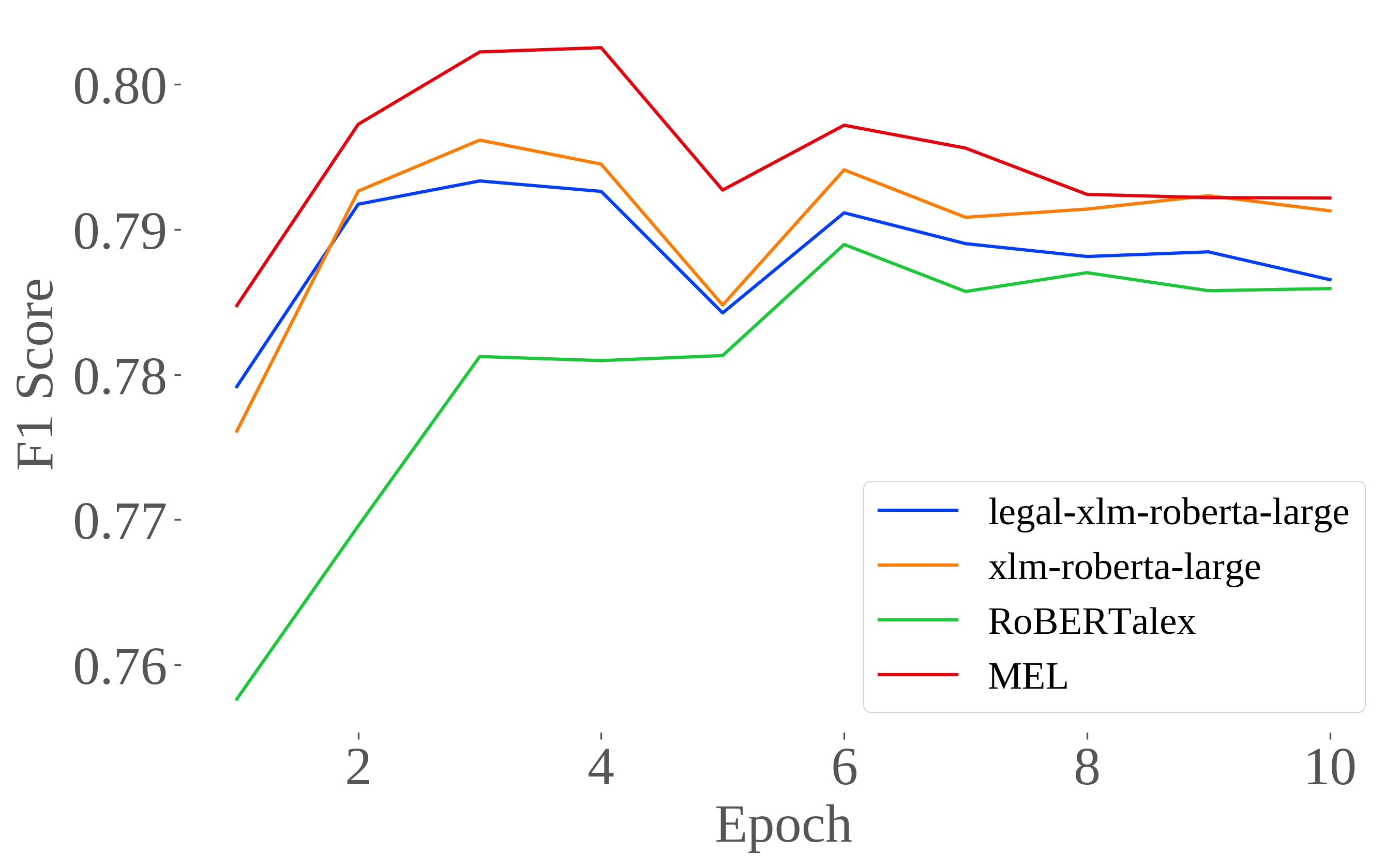}
    \caption{Multieurlex F1 vs Epochs}
    \label{fig:results_eurlex_image}
  \end{minipage}
  \hfill
  \begin{minipage}[b]{0.49\textwidth}
    \centering
    \resizebox{\textwidth}{!}{%
      \begin{tabular}{lcc}
      \toprule
      \textbf{Model} & \textbf{Max F1} & \textbf{Epochs vs F1 AUC} \\
      \midrule
      Legal-XLM-RoBERTa-Large & 0.7933 & 7.1016 \\
      XLM-RoBERTa-Large       & 0.7962 & 7.1205 \\
      RoBERTalex              & 0.7890 & 7.0324 \\
      MEL                     & \textbf{0.8025} & \textbf{7.1606} \\
      \bottomrule
      \end{tabular}
    }
    \captionof{table}{Max F1 scores and learning speed for various models on the Multieurlex dataset. Bold values indicate the best scores.}
    \label{table:results_eurlex_table}
  \end{minipage}
\end{figure}

\item \textbf{Private Multiclass dataset}:
Similar to the evaluation on the eurlex dataset, MEL seems to begin the finetuning with a lot of alredy-known information, achieving higher results after 1 epoch than the RoBERTalex model after a complete finetune. MEL results in higher F1 and learning speed than the rest of the models.

\begin{figure}[ht]
  \begin{minipage}[b]{0.49\textwidth}
    \includegraphics[width=\textwidth]{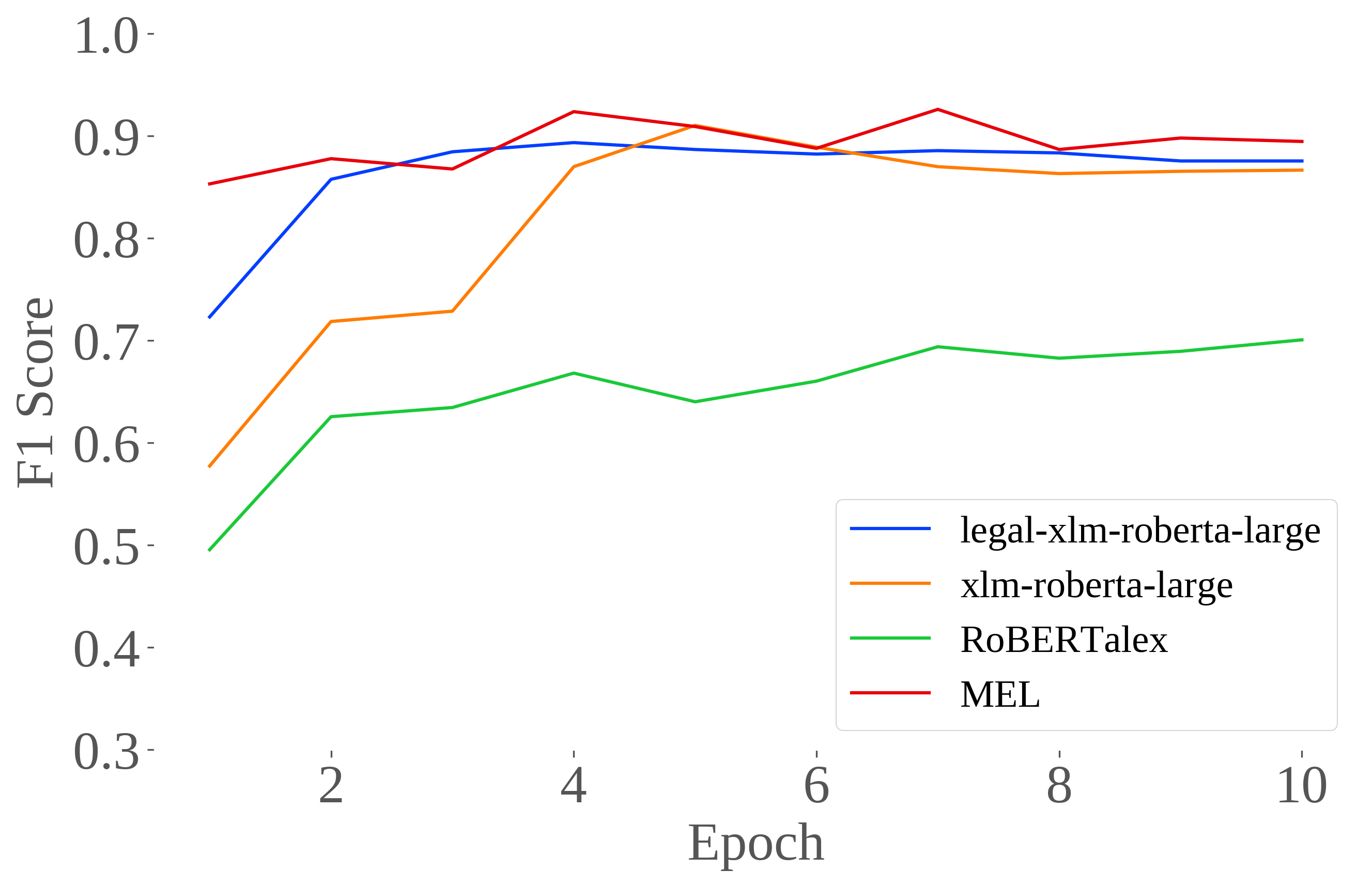}
    \caption{Multiclass F1 vs Epochs}
    \label{fig:legal_classification_results_image}
  \end{minipage}
  \hfill
  \begin{minipage}[b]{0.49\textwidth}
    \centering
    \resizebox{\textwidth}{!}{%
      \begin{tabular}{lcc}
      \toprule
      \textbf{Model} & \textbf{Max F1} & \textbf{Epochs vs F1 AUC} \\
      \midrule
      Legal-XLM-RoBERTa-Large & 0.8935 & 7.8487 \\
      XLM-RoBERTa-Large       & 0.9103 & 7.4372 \\
      RoBERTalex              & 0.7007 & 5.8929 \\
      MEL                     & \textbf{0.9260} & \textbf{8.0510} \\
      \bottomrule
      \end{tabular}
    }
    \captionof{table}{Max F1 scores and learning speed for various models on the Multieurlex dataset. Bold values indicate the best scores.}
    \label{table:legal_classification_results_table}
  \end{minipage}
\end{figure}

Following the results on Figure \ref{fig:legal_classification_results_image}, some additional experiments were performed to analyze the behavior of the models on small data. To achieve this, steps previous to the first epoch were saved to explore how the model would perform when training on fewer data. Results can be observed on Figure \ref{fig:small_data_image} and Table \ref{table:small_data_table}. These results confirm that MEL is better aligned with the legal language, since with small data it is already able to achieve a significantly better F1 than the other models.

\begin{figure}[ht]
  \begin{minipage}[b]{0.49\textwidth}
    \includegraphics[width=\textwidth]{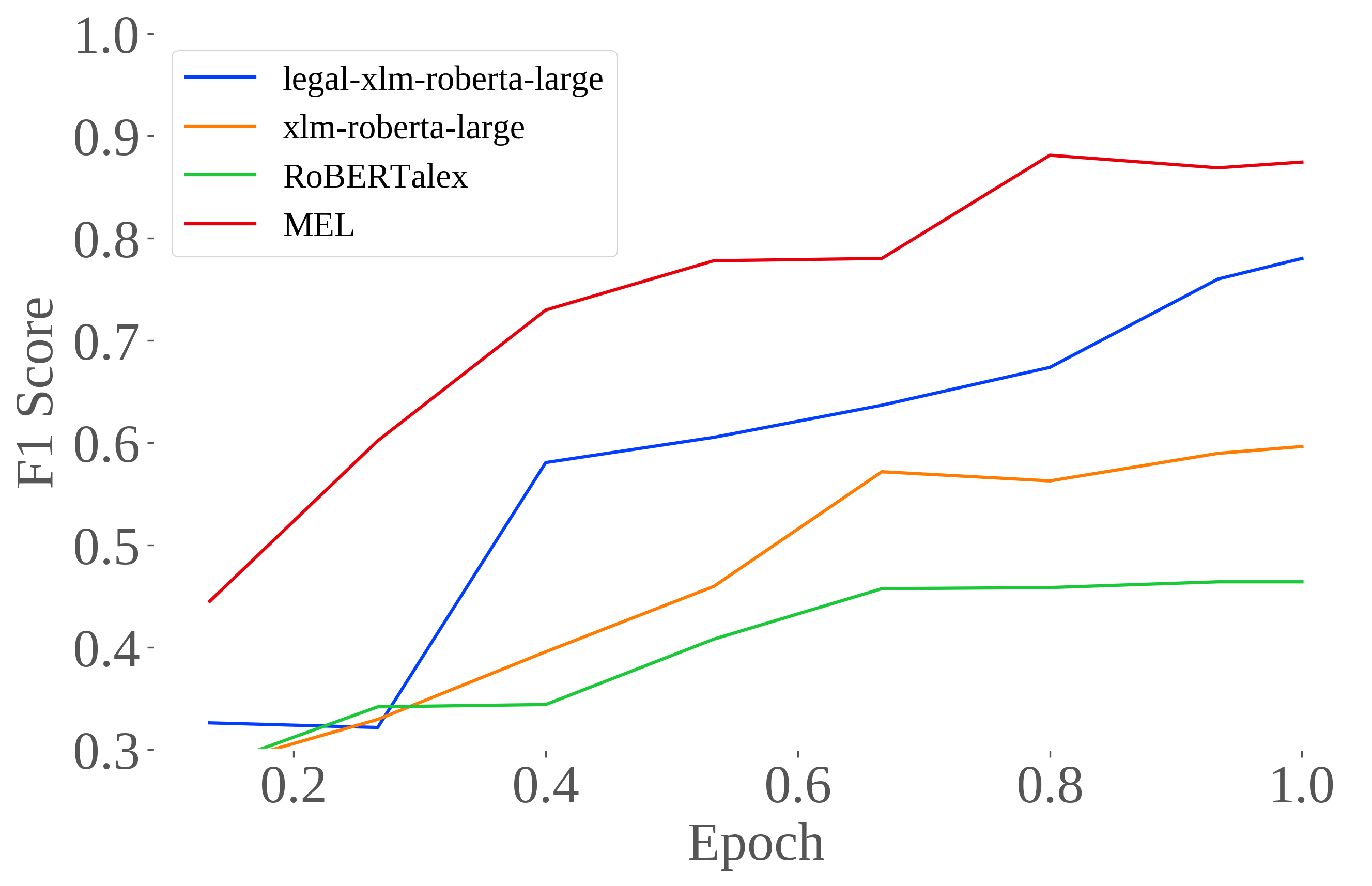}
    \caption{Multiclass F1 vs Epochs (small data)}
    \label{fig:small_data_image}
  \end{minipage}
  \hfill
  \begin{minipage}[b]{0.49\textwidth}
    \centering
    \resizebox{\textwidth}{!}{%
      \begin{tabular}{lcc}
      \toprule
      \textbf{Model} & \textbf{Max F1} & \textbf{Epochs vs F1 AUC} \\
      \midrule
      Legal-XLM-RoBERTa-Large & 0.7803 & 0.4946 \\
      XLM-RoBERTa-Large       & 0.5964 & 0.4070 \\
      RoBERTalex              & 0.4641 & 0.3487 \\
      MEL                     & \textbf{0.8812} & \textbf{0.6485} \\
      \bottomrule
      \end{tabular}
    }
    \captionof{table}{Max F1 scores and learning speed for various models on the Multieurlex dataset. Bold values indicate the best scores.}
    \label{table:small_data_table}
  \end{minipage}
\end{figure}

\end{itemize}

\section{Discussion and Conclusion}
The results of this study highlight the importance of domain-specific pre-training for legal text classification, particularly in a language as resource-scarce as Spanish for legal corpora. By comparing our pretrained model, MEL, against both general-purpose multilingual models and existing domain-specific models, several key observations emerge.

\begin{itemize}
\item \textbf{Importance of Domain-Specific Pre-training}: The superior performance of MEL across both multilabel and multiclass classification tasks underscores the value of domain-specific pre-training. While general-purpose models like xlm-roberta-large can capture linguistic features across multiple languages, they lack the depth required for complex and technical vocabulary aspects present in legal texts. This observation aligns with findings in other domains, where pre-training on domain-specific corpora has proven to yield better results on task-specific evaluations.

\item \textbf{Comparison to Other Domain-Specific Models}: While joelniklaus/legal-xlm-roberta-large is also a domain-specific model, its performance was consistently outpaced by MEL. This difference could be attributed to several aspects:
\begin{itemize}
    \item \textbf{Focus on Spanish Legal Texts}: Given that the purpose of our model was to achieve good results in Spanish, the focus on Spanish texts allowed the model to better understand language-specific features that the multilingual model may not capture with rigorous detail.
    \item \textbf{Pre-training Corpus Quality}: By narrowing the corpus to Spanish language, a more meticulous cleaning was performed, providing a higher-quality dataset and allowing the model to learn from less corrupted data or noise.
\end{itemize}

\item \textbf{Challenges in Spanish Legal Tasks}: Token or span classification tasks remain unevaluated due to the lack of annotated texts. This highlights the need for further development of labeled datasets in the legal domain.
\end{itemize}

\subsection{Future Work}
Future work could involve several directions to address the limitations and extend the applicability of this study:
\begin{itemize}
\item Developing annotated datasets for token or span classification tasks in Spanish legal texts to facilitate evaluation and fine-tuning.
\item Expanding the pre-training corpus to include complex legal text extraction tasks, enabling models like MEL to handle more nuanced legal applications.
\item Testing MEL on diverse downstream tasks such as question answering, information retrieval, and summarization within the legal domain.
\end{itemize}

\textbf{Acknowledgements}

This work has received funding from the INESData project (Infrastructure to Investigate Data Spaces in Distributed Environments at UPM), a project funded under the UNICO I+D CLOUD call by the Ministry for Digital Transformation and the Civil Service, in the framework of the recovery plan PRTR financed by the European Union (NextGenerationEU).

\bibliographystyle{unsrt}  
\bibliography{references}  

\begin{thebibliography}{10}

\bibitem{vaswani2023attentionneed}
Ashish Vaswani, Noam Shazeer, Niki Parmar, Jakob Uszkoreit, Llion Jones, Aidan~N. Gomez, Lukasz Kaiser, and Illia Polosukhin.
\newblock Attention is all you need, 2023.

\bibitem{bert}
Jacob Devlin, Ming-Wei Chang, Kenton Lee, and Kristina Toutanova.
\newblock Bert: Pre-training of deep bidirectional transformers for language understanding, 2019.

\bibitem{xlmroberta}
Alexis Conneau, Kartikay Khandelwal, Naman Goyal, Vishrav Chaudhary, Guillaume Wenzek, Francisco Guzmán, Edouard Grave, Myle Ott, Luke Zettlemoyer, and Veselin Stoyanov.
\newblock Unsupervised cross-lingual representation learning at scale, 2020.

\bibitem{wu2019betobentzbecassurprising}
Shijie Wu and Mark Dredze.
\newblock Beto, bentz, becas: The surprising cross-lingual effectiveness of bert, 2019.

\bibitem{niklaus2024multilegalpile689gbmultilinguallegal}
Joel Niklaus, Veton Matoshi, Matthias Stürmer, Ilias Chalkidis, and Daniel~E. Ho.
\newblock Multilegalpile: A 689gb multilingual legal corpus, 2024.

\bibitem{huggingfaceJoelniklauslegalxlmrobertalargeHugging}
joelniklaus/legal-xlm-roberta-large · {H}ugging {F}ace --- huggingface.co.
\newblock \url{https://huggingface.co/joelniklaus/legal-xlm-roberta-large}.
\newblock [Accessed 22-11-2024].

\bibitem{warner2024smarterbetterfasterlonger}
Benjamin Warner, Antoine Chaffin, Benjamin Clavié, Orion Weller, Oskar Hallström, Said Taghadouini, Alexis Gallagher, Raja Biswas, Faisal Ladhak, Tom Aarsen, Nathan Cooper, Griffin Adams, Jeremy Howard, and Iacopo Poli.
\newblock Smarter, better, faster, longer: A modern bidirectional encoder for fast, memory efficient, and long context finetuning and inference, 2024.

\bibitem{subies2023surveyspanishclinicallanguage}
Guillem~García Subies, Álvaro Barbero~Jiménez, and Paloma~Martínez Fernández.
\newblock A survey of spanish clinical language models, 2023.

\bibitem{biobert}
Jinhyuk Lee, Wonjin Yoon, Sungdong Kim, Donghyeon Kim, Sunkyu Kim, Chan~Ho So, and Jaewoo Kang.
\newblock Biobert: a pre-trained biomedical language representation model for biomedical text mining.
\newblock {\em Bioinformatics}, 36(4):1234–1240, September 2019.

\bibitem{scibert}
Iz~Beltagy, Kyle Lo, and Arman Cohan.
\newblock Scibert: A pretrained language model for scientific text, 2019.

\bibitem{chalkidis-etal-2020-legal}
Ilias Chalkidis, Manos Fergadiotis, Prodromos Malakasiotis, Nikolaos Aletras, and Ion Androutsopoulos.
\newblock {LEGAL}-{BERT}: The muppets straight out of law school.
\newblock In {\em Findings of the Association for Computational Linguistics: EMNLP 2020}, pages 2898--2904, Online, November 2020. Association for Computational Linguistics.

\bibitem{caselawbert}
Shounak Paul, Arpan Mandal, Pawan Goyal, and Saptarshi Ghosh.
\newblock Pre-trained language models for the legal domain: A case study on indian law, 2023.

\bibitem{PLN6487}
Rodrigo~Agerri y~Eneko~Agirre.
\newblock Lessons learned from the evaluation of spanish language models.
\newblock {\em Procesamiento del Lenguaje Natural}, 70(0):157--170, 2023.

\bibitem{joelniklaus2023legalxlmr}
Joel Niklaus.
\newblock legal-xlm-roberta-large.
\newblock \url{https://huggingface.co/joelniklaus/legal-xlm-roberta-large}, 2023.
\newblock Multilingual legal language model pretrained on Multi Legal Pile.

\bibitem{Niklaus2023MultiLegalPileA6}
Joel Niklaus, Veton Matoshi, Matthias Sturmer, Ilias Chalkidis, and Daniel~E. Ho.
\newblock Multilegalpile: A 689gb multilingual legal corpus.
\newblock {\em ArXiv}, abs/2306.02069, 2023.

\bibitem{datos_congreso}
Jesus Cerquides~Bueno and Paula Mateos~Marín.
\newblock Spanish congress parliamentary records (1977-2024), 2024.

\bibitem{pdfdirective}
Council of the European~Union European~Parliament.
\newblock Directive (eu) 2016/2102 of the european parliament and of the council of 26 october 2016 on the accessibility of the websites and mobile applications of public sector bodies (text with eea relevance).
\newblock \url{http://data.europa.eu/eli/dir/2016/2102/oj}, 2016.

\bibitem{Singer-Vine_pdfplumber_2024}
Jeremy Singer-Vine and {The pdfplumber contributors}.
\newblock {pdfplumber}, August 2024.

\bibitem{Facebookresearch}
Facebookresearch.
\newblock Facebookresearch/fasttext: Library for fast text representation and classification.

\bibitem{huggingfaceTrainer}
{T}rainer --- huggingface.co.
\newblock \url{https://huggingface.co/docs/transformers/main_classes/trainer}.
\newblock [Accessed 20-12-2024].

\bibitem{chalkidis2021multieurlexmultilingualmultilabel}
Ilias Chalkidis, Manos Fergadiotis, and Ion Androutsopoulos.
\newblock Multieurlex -- a multi-lingual and multi-label legal document classification dataset for zero-shot cross-lingual transfer, 2021.

\bibitem{huggingfacePlanTLGOBESRoBERTalexHugging}
{P}lan{T}{L}-{G}{O}{B}-{E}{S}/{R}o{B}{E}{R}{T}alex · {H}ugging {F}ace --- huggingface.co.
\newblock \url{https://huggingface.co/PlanTL-GOB-ES/RoBERTalex}.
\newblock [Accessed 22-11-2024].

\end{thebibliography}






\end{document}